\documentclass[10pt,twocolumn,letterpaper]{article}

\usepackage{iccv}
\usepackage{times}
\usepackage{epsfig}
\usepackage{graphicx}
\usepackage{amsmath}
\usepackage{amssymb}
\usepackage{multirow}
\usepackage{subfigure}
\usepackage{url}
\usepackage{color}
\usepackage{balance}
\usepackage[breaklinks=true,bookmarks=false]{hyperref}

\iccvfinalcopy % *** Uncomment this line for the final submission

\def\iccvPaperID{4142} % *** Enter the ICCV Paper ID here

\ificcvfinal\fi

\begin{document}

%%%%%%%%% TITLE
\title{From General to Specific: Informative Scene Graph Generation via Balance Adjustment}

\author{Yuyu Guo\textsuperscript{1} 
	\and Lianli Gao\textsuperscript{1}\thanks{Corresponding author.} 
	\and Xuanhan Wang\textsuperscript{1} 
	\and Yuxuan Hu\textsuperscript{2} 
	\and Xing Xu\textsuperscript{1} 
	\and Xu Lu\textsuperscript{3}
	\and Heng Tao Shen\textsuperscript{1} 
	\and Jingkuan Song\textsuperscript{1}  \and
\textsuperscript{1}Center for Future Media \& School of Computer Science and Engineering, \\ University of Electronic Science and Technology of China, China \and \textsuperscript{2}Southwest University, China \and \textsuperscript{3}Kuaishou, China}

\maketitle
%\thispagestyle{empty}
% Remove page # from the first page of camera-ready.
\ificcvfinal\thispagestyle{empty}\fi
%%%%%%%%% ABSTRACT
\begin{abstract}  
	The scene graph generation (SGG) task aims to detect visual relationship triplets, i.e., subject, predicate, object, in an image, providing a structural vision layout for scene understanding. However, current models are stuck in common predicates, e.g., ``on'' and ``at'', rather than informative ones, e.g., ``standing on'' and ``looking at'', resulting in the loss of precise information and overall performance. If a model only uses ``stone on road'' rather than ``blocking'' to describe an image, it is easy to misunderstand the scene. We argue that this phenomenon is caused by two key imbalances between informative predicates and common ones, i.e., \textbf{semantic space level imbalance} and \textbf{training sample level imbalance}. To tackle this problem, we propose BA-SGG, a simple yet effective SGG framework based on \textbf{balance adjustment} but not the conventional distribution fitting. It integrates two components: \textbf{Semantic Adjustment (SA)} and \textbf{Balanced Predicate Learning (BPL)}, respectively for adjusting these imbalances. Benefited from the model-agnostic process, our method is easily applied to the state-of-the-art SGG models and significantly improves the SGG performance. Our method achieves 14.3\%, 8.0\%, and 6.1\% higher Mean Recall (mR) than that of the Transformer model at three scene graph generation sub-tasks on Visual Genome, respectively. Codes are publicly available\footnote{\url{https://github.com/ZhuGeKongKong/SGG-G2S}}.
\end{abstract}

%%%%%%%%% BODY TEXT

\section{Introduction}
\label{sec:intro}
\begin{figure}[t]
	\begin{center}	
		
		\subfigure[]
		{
			\includegraphics[width=0.9\linewidth]{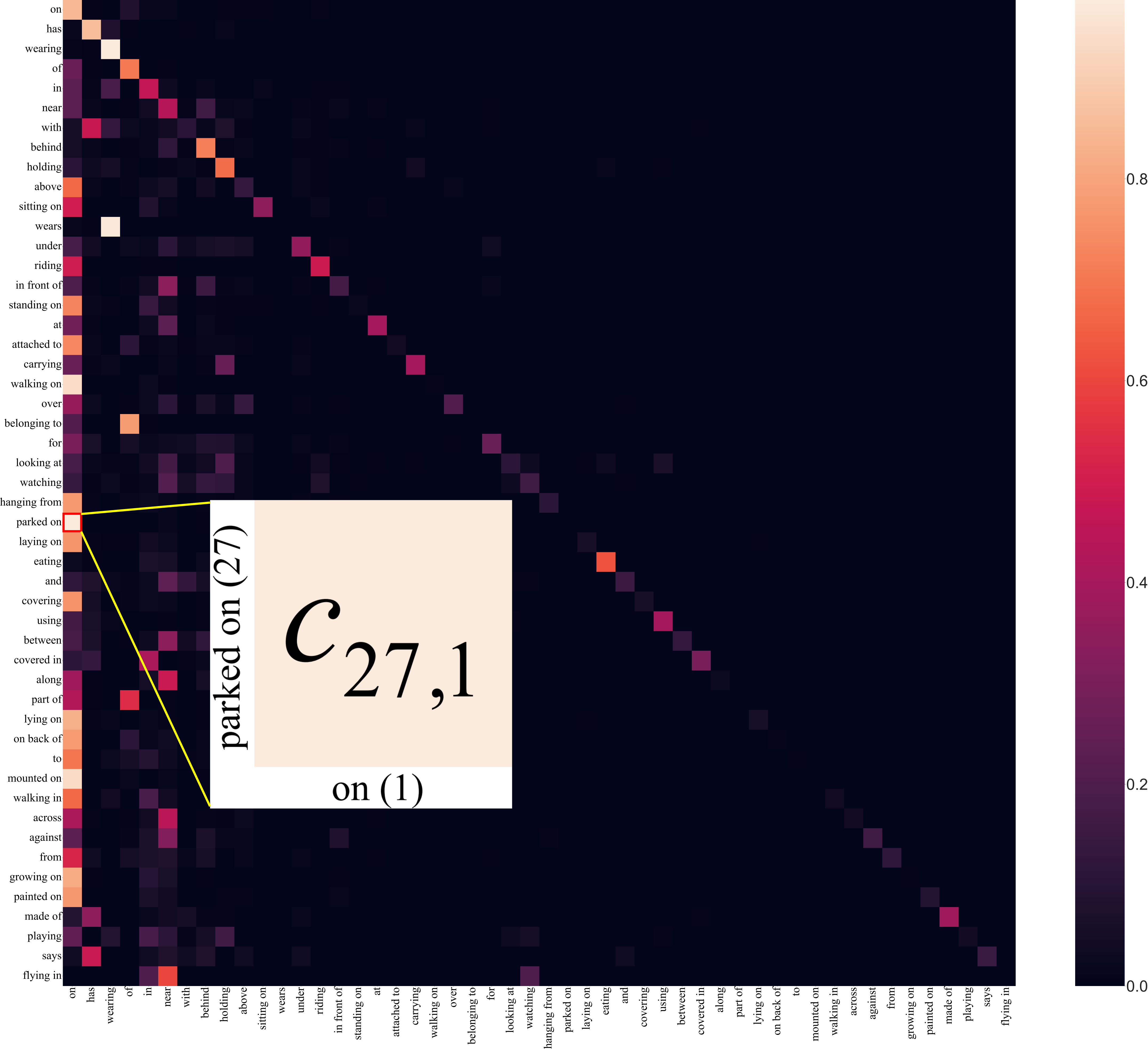} 
			\label{fig:conf_mat1}
		}
 
	    \vspace{-0.7em}
		\subfigure[]
		{
			\includegraphics[width=0.9\linewidth]{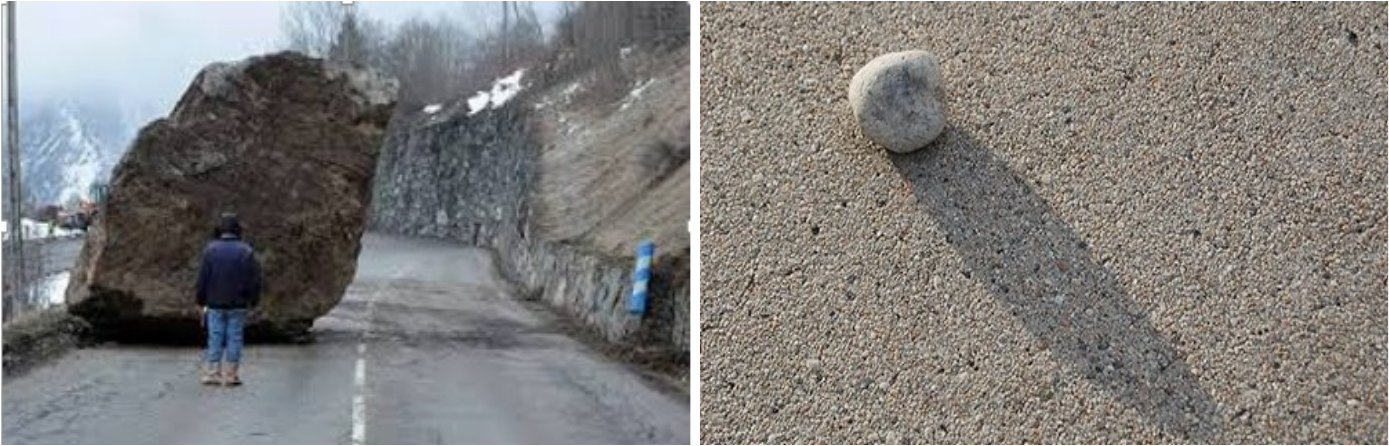}  	
			\label{fig:conf_mat2}
		}
    \vspace{-1.0em}
	\end{center}
	\caption{A confusion matrix of a baseline model (a) and two searching results for ``stone on road'' in Google (b). The element $c_{i,j}$ means the number of samples labeled as $i$ but predicted as $j$. The results of the model concentrate on the left side, \eg, ``on'' and ``near''. If the model directly uses the common predicate ``on'' to describe the first image like the second image in (b), it may cause serious consequences.} 
	\label{fig:conf_mat}
	\vspace{-1.5em}
\end{figure}
Scene Graph Generation (SGG) aims to detect instances-of-interest and their relationships in an image. It provides a structural vision layout, as an auxiliary tool, to bridge the gap from computer vision to natural language, supporting many high-level tasks such as visual captioning~\cite{img_cap:attri,img_cap:bottomup} and visual question answering~\cite{vqa:zp1,vqa:xp1}. 

A good scene graph can provide sufficient and informative relationships among instance-of-interest. Inspired by remarkable progress in object detection~\cite{obj_det:rcnn,obj_det:fastrcnn,obj_det:yolo}, existing solutions~\cite{scenegraph:GPI,scenegraph:IMP,scenegraph:factnet,scenegraph:zoomnet} mostly follow a common generation paradigm, that is, detecting objects from an image, extracting region features, and then recognizing the predicate categories under the guidance of standard classification objective function. Based on this paradigm, current state-of-the-art methods, however, exhibit the tendency that most of the recognized predicates are general or common (\eg, ``on'' or ``in'') and lack specific or informative contents (\eg, ``standing on'' or ``sitting on''), as shown in Fig.~\ref{fig:conf_mat1}. Thus, it is hard to apply an SGG algorithm in real-world settings like Fig.~\ref{fig:conf_mat2}, since it provides insufficient clues and may lead to misunderstanding about a scene. In this work, our study reveals that such biased predicates are caused by imbalance issues, preventing the power of the well-designed SGG model from being exploited. Specifically, this issue can further be divided into two sub-problems: \textit{semantic space level imbalance} and \textit{training sample level imbalance}.

\begin{figure}[t]
	\begin{center}
		\includegraphics[width=1.0\linewidth]{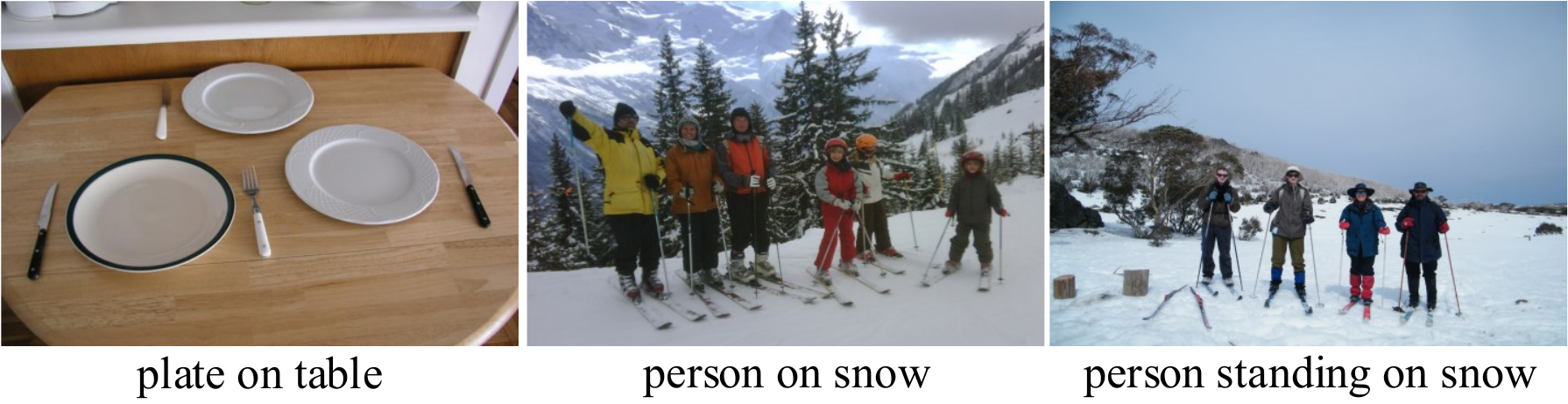}
	\end{center}
\vspace{-1.0em}
	\caption{Ground-truth annotations from Visual Genome. The first and second examples are marked as ``on'' because of the huge semantic space of ``on''. Compared with ``person standing on snow'' in the third example, ``person on snow'' in the second example is ambiguous and information-poor.}
	\label{fig:conf_ann}
	\vspace{-1.5em}
\end{figure}

\noindent \textbf{\textit{Semantic space level imbalance.}} Common predicates, such as ``on'', have large semantic spaces, while informative predicates like ``standing on'' provide rich content but have small semantic spaces and even may be replaced by common ones in some annotations. Some examples of this semantic space level imbalance are shown in Fig.~\ref{fig:conf_ann}. All examples in Fig.~\ref{fig:conf_ann} reflect the fact of ``on'', while only the last two examples represent identical contents of ``standing on''. Although ``standing on'' is more specific for the last two examples, humans still label the second image with ``on'' because of its larger semantic space than ``standing on''. The semantic space level imbalance also reflects the semantic relationship between predicates, \eg, annotators prefer to label ``standing on'' as ``on'' rather than ``has''. Tagging different labels for identical contents confuses the SGG model and causes poor performance.

\begin{figure}[t]
	\begin{center}
		\includegraphics[width=0.95\linewidth]{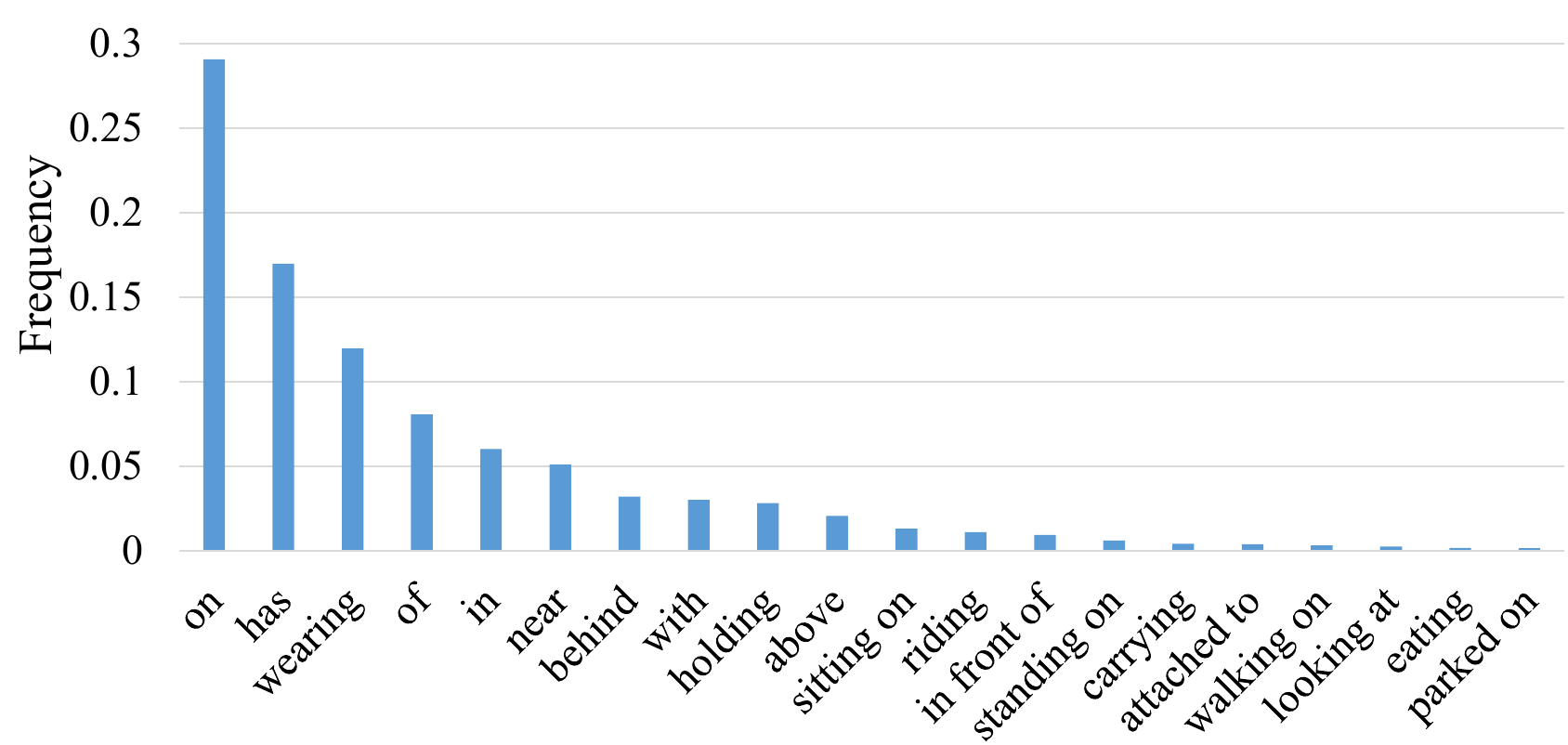}
	\end{center}
\vspace{-1.0em}
	\caption{Frequencies of predicates in the Visual Genome dataset. For clarity, we omit some predicates.}
	\label{fig:pred_dist}
	\vspace{-1.5em}
\end{figure}

\noindent\textbf{\textit{Training sample level imbalance.}} When learning to recognize predicates, informative predicate samples are particularly valuable as they provide precise knowledge to a scene graph. However, predicate samples are dominated by the common predicate categories in the SGG dataset (\eg, Visual Genome). For example, Fig.~\ref{fig:pred_dist} shows the annotation distribution over predicate categories. We can observe that the number of common predicate samples is much larger than that of informative ones, which leads to the problem of the long-tailed distribution of the classes. Within such an imbalanced sampling space, the prediction of informative predicates is dominated by the common ones. 

The above challenges motivate us to study two problems: 1) how to revise common predicates as informative ones based on the semantic relation. And 2) how to make training sample space balanced.
To tackle these problems, we believe that the key is to explore the predicate relation in semantic space and adjust sampling space to be balanced. Motivated by this, we propose a simple yet effective pipeline, namely, \textit{Scene Graph Generation with Balance Adjustment} (BA-SGG), to learn informative scene graph generation. This pipeline integrates two novel components: 1) Semantic Adjustment (SA), which casts common predictions as informative ones with rich information contents by fully exploiting semantic relation among predicates. 2) Balanced Predicate Learning (BPL), which mines informative predicates according to their information contents.

To sum up, the main contributions of our work contain: 1). We systematically review the process of scene graph generation and reveal the problem of insufficient information contents that limit SGG's overall performance and practical applicability, \ie, \textit{semantic space level imbalance} and \textit{training sample level imbalance}; 2). We propose the BA-SGG for informative scene graph generation, a novel framework that adjusts biased predicate predictions by two components: semantic adjustment (SA) and balanced predicate learning (BPL); 3). The proposed BA-SGG is tested on Visual Genome, obtaining significant improvements over state-of-the-art SGG approaches. Without bells and whistles, BA-SGG achieves 14.3\%, 8.0\%, and 6.1\% higher Mean Recall (mR) than Transformer~\cite{networks:transformer,scenegraph:sggcode} on three scene graph generation sub-tasks, respectively. Besides, we propose a new metric (mRIC@K) to measure the information content of results.

%-------------------------------------------------------------------------

\section{Related Work}
\noindent\textbf{Scene Graph Generation.} Many methods~\cite{scenegraph:GPI,scenegraph:IMP,scenegraph:factnet,scenegraph:zoomnet} were proposed to handle the scene graph generation task in recent years. These methods deal with the task from different perspectives, such as extracting context information by message passing~\cite{scenegraph:IMP,scenegraph:motifs,scenegraph:hga,scenegraph:gps}, constructing visual embedding in the semantic space~\cite{scenegraph:VTEN,scenegraph:cte,scenegraph:PGAE}, improving robustness by external knowledge~\cite{scenegraph:kggan,scenegraph:one-shot,scenegraph:VRD_LP}. 
Except for these models trapped in common predicates, Liang~\cite{scenegraph:vrg_iccv} reconstructed a dataset that focuses on visually-relevant relationships. The unbiased SGG is proposed in~\cite{scenegraph:ubtraing} to remove the vision-agnostic bias with counterfactual causality. Different from these works, our method explores the imbalance in the semantic and learning space adequately to learn scene graphs with precise and rich information.

\noindent\textbf{Informative Prediction.} In order to get out of common prediction, many works have been proposed in computer vision, such as fine-grained recognition~\cite{fg:chen,fg:mic,fg:wang} and informative image captioning~\cite{inf_cap:reu,inf_cap:san}. Most of them focus on the informative object categories, which are either precisely represented by samples or constructed with distinct hierarchies. Unlike object categories, it is difficult to define a clear hierarchical or coarse-to-fine structure for relationship predicates. Instead of seeking a way to define a structure among predicates manually, we estimate a semantic relation from a baseline model and define the predicate information content to learn informative predicates.

\section{Approach}
\subsection{Approach Overview}
\label{sec.proform}

\begin{figure}[t]
	\begin{center}
		\includegraphics[width=0.9\linewidth]{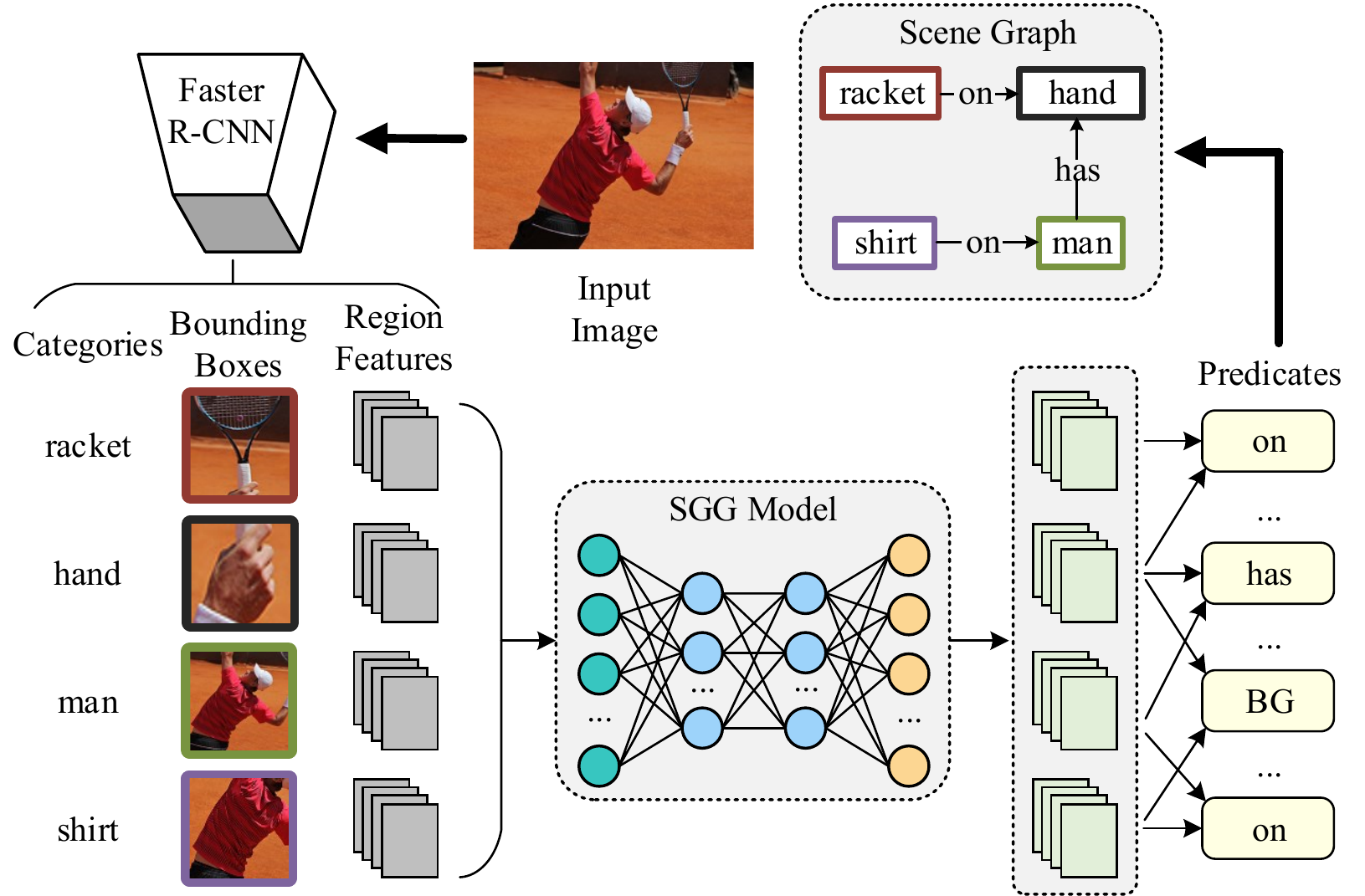}
	\end{center}
\vspace{-1.0em}
	\caption{The main process of previous methods from the input image to the scene graph.}
	\label{fig:gsgg_framework}
	\vspace{-1.5em}
\end{figure}
\begin{figure}[t]
	\begin{center}
		\includegraphics[width=0.8\linewidth]{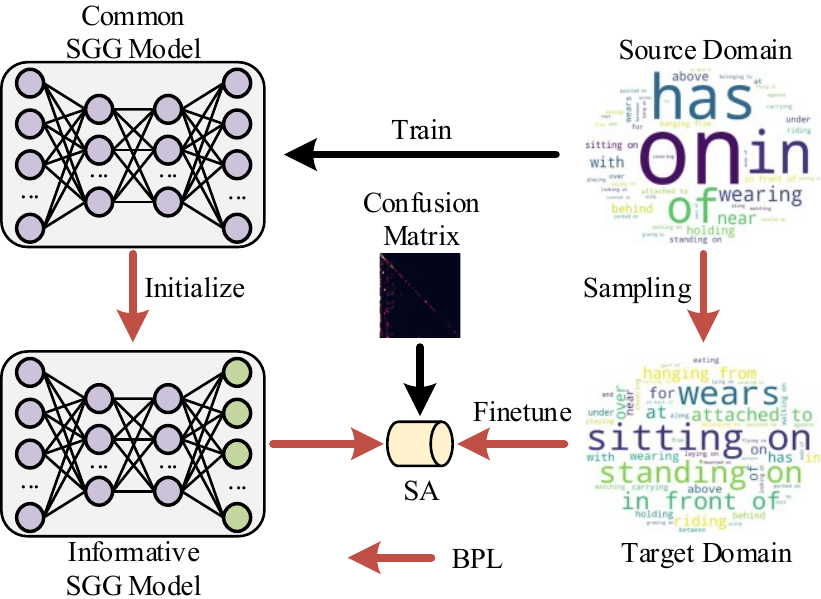}
	\end{center}
	\vspace{-1.0em}
	\caption{The whole training process of our method, \ie, Semantic Adjustment (SA) and Balanced Predicate Learning (BPL).}
	\label{fig:g2s_framework}
	\vspace{-1.5em}
\end{figure}

\noindent\textbf{Problem Formulation.} As shown in Fig.~\ref{fig:gsgg_framework}, conventional methods formulate scene graph generation as a two-stage process, where it firstly detects all instances-of-interest and then recognizes relationship predicates between pair-wise instances. 
Given an image $\mathbf{X}$, its corresponding scene graph is generated from the complete graph $\mathbf{G} = <\mathbf{O}, \mathbf{R}>$, where $\mathbf{O}$ is the set of instance nodes. $\mathbf{R}$ is the full set of edges, each of which connects two nodes and encodes a relationship between them. More specifically, each edge is represented in the form of a triplet $( o_i, y_{ij}, o_j )$, where $o_i$ denotes the subject, $o_j$ is the object, and $y_{ij}$ represents the predicate of the edge/triplet. We denote by $\mathbf{Y}$ the list of predicate labels for the edges $\mathbf{R}$, hence $|\mathbf{Y}|=|\mathbf{R}|=|\mathbf{O}\times\mathbf{O}|$, where $|\cdot|$ represents the length/size of a list/set. The pair of $(o_i, o_j)$ is embodied as region bounding boxes with object categories, which are usually obtained by an object detector such as Faster R-CNN~\cite{obj_det:faster}. Thus, SGG models omit the process of object detection and only focus on predicting relationship predicates $\mathbf{Y}$ between instance pairs, by maximizing the probability $\Pr ({\mathbf{Y}}|{\mathbf{O}}\times\mathbf{O}; \theta)$, where $\theta$ is the learnable parameter of the scene graph generation model. 

\noindent\textbf{Overview.} We follow this pipeline approach and focus on the second stage, \ie, predicate recognition, after instance regions are detected by an object detection system. In particular, the proposed method consists of two processes, as shown in Fig.~\ref{fig:g2s_framework}. 1) Semantic Adjustment (SA) is applied to casting common predictions generated by an SGG model as informative ones, where it performs a relation modeling among predicates. 2) Balanced Predicate Learning (BPL) is designed for extending the sampling space for informative predicates. In the following subsections, we present details of each component.

\subsection{Semantic Adjustment}
As we discussed earlier, most of the current SGG models bias toward the common predicate classes. The Semantic Adjustment component is designed to restore informative predictions from common ones, based on the semantic relationship between predicates. Intuitively, ``standing on'' is more likely to be predicted as ``on'' than ``has'' because the meanings of ``standing on'' and ``on'' are closer than between ``standing on'' and ``has''. Therefore, we exploit such relationships to adjust the prediction results.
In particular, we formalize this adjustment as Eq.~\ref{equ.transform}: 
\begin{equation}
	\label{equ.transform}
	\begin{aligned}
		%\Pr(\mathbf{Y}|\mathbf{O}\times\mathbf{O};\theta)&=\Pi_{o_i,o_j\in\mathbf{O}}\Pr(y_{ij}^s|o_i,o_j;\theta)\\
		\Pr(y_{ij}^s | o_i, o_j;\theta) &= \Pr(y_{ij}^s | y_{ij}^g) \Pr( y_{ij}^g | o_i, o_j;\theta )~,
	\end{aligned}
\end{equation}
where $\Pr( y_{ij}^g | o_i, o_j;\theta ) \in \mathbb{R}^{K}$ is the prediction of an SGG model for all predicate categories between subject $o_i$ and object $o_j$, $\Pr({y_{ij}^s} | o_i, o_j;\theta) \in \mathbb{R}^{K}$ is produced by the SGG model after semantic adjustment, and $K$ is the total number of predicate categories. The superscripts $g$ and $s$ are used to distinguish the prediction before and after SA. $\Pr(y_{ij}^s | y_{ij}^g)$ represents the semantic adjustment, which measures how confident a common result is transformed into an informative one. 

The semantic adjustment $\Pr(y^s_{ij}|y^g_{ij})$ reflects the latent semantic relationship between predicates. It can be considered as a transition matrix between pairs of predicate labels, thus is of dimension $K\times K$. 
For simplicity, we denote $\Pr(y_{ij}^s | y_{ij}^g)$ as a transition matrix $C^{\star} \in \mathbb{R}^{K \times K}$, and it can be derived from prediction priors. We denote by $C \in \mathbb{R}^{K \times K}$ the prediction confusion matrix, where each element $c_{kl}$ denotes the number of samples labeled as the $k$-th category but being predicted as $l$. In this paper, we use the frequency model~\cite{scenegraph:motifs} to generate predictions on the training dataset for $C$, and fix it in our framework. Next, a normalized transition matrix $C'$, in which each element $c'_{kl}$ is defined in Eq.~\ref{equ.norm} as follows:
\begin{equation}
	\label{equ.norm}
	\begin{aligned}
		c'_{kl} {\rm{ = }}\frac{{{c_{kl}}}}{{\sum\limits_{m = 1}^K {{c_{km}}} }}~.
	\end{aligned}
\end{equation}
The visualization of the row-normalized confusion matrix is shown in the heatmap in Fig.~\ref{fig:conf_mat1}. We can observe that large values of the matrix concentrate on the left side. This implies that many informative predicates are classified as a few common ones. The matrix also implies the semantic relationship between predicates to some extent, \eg, ``parked on'' is more likely to be predicted ``on'' than ``has''. 
However, if an informative predicate has been predicted, its probability should not be adjusted significantly. Directly multiplying $C'$ may greatly decrease the score of an informative predicate because the diagonal tail values are small.
Therefore, the diagonal elements' values should be increased to alleviate this impact.
To achieve this goal, we add a weighted identity matrix to $C'$, resulting in the final transition matrix $C^{\star}$ as formed by Eq.~\ref{equ.final_transition}:
\begin{equation}
	\label{equ.final_transition}
	\begin{aligned}
		C^{\star} = Row\_Normalize(C'+ \alpha I_K)~,
	\end{aligned}
\end{equation}
where $I_K\in \mathbb{R}^{K \times K}$ is the identity matrix and $\alpha$ is a hyper-parameter. Normalizing the matrix by row ensures that the sum of the row is $1$. We freeze the transition probability matrix during training to avoid semantic drifting.

\subsection{Balanced Predicate Learning}
\label{sec.pdtl}
To address imbalanced training, a new data source with a balanced sample space is needed. Formally, we view a data source with an imbalanced sample space as a source domain, and a data source with a relatively balanced sample space as a target domain. Unlike existing methods that train the SGG model in the source domain only, balanced predicate learning (BPL) divides the learning process into three stages: 1). Training SGG model on a source domain; 2). Creating a target domain; 3). Transfer learning on the target domain. Fig.~\ref{fig:g2s_framework} shows the overview of this learning process.

In the first stage, we take a common training strategy that is identical to that of previous works~\cite{scenegraph:motifs,scenegraph:IMP} to train an SSG model on the source domain.

In the second stage, a target domain with a relatively balanced sample space is constructed. However, it is hard to decide whether a predicate is informative or not, since annotations related to the predicate property are not provided. Intuitively, common predicates are semantically general, while informative predicates are context-specific. Therefore, that discriminative problem can be transformed into a process of measurement, where it measures how informative an event is, in other words, how much information content does an event involves. According to Shannon information theory, the information content $\mathrm{I}$ of an event $z$ can be assessed by Eq.~\ref{equ.shanno}:
\begin{equation}
	\label{equ.shanno}
	\begin{aligned}
		\mathrm{I}(z) =-\log _{b}[\Pr(z)]~,
	\end{aligned}
\end{equation}
where $\Pr(z)$ denotes the probability of the event $z$. It reflects the fact that an event with a small probability provides more information content. It is noteworthy that other ways of calculating `informativeness', \eg, semantic context based method~\cite{LiangWHG15} or statistical information in corpora by TF-IDF~\cite{infor:ifidf}, can also be used here. We calculate `informativeness' based on Shannon information theory mainly for simplicity and general cases. 
Inspired by this, we measure informativeness for each predicate category by estimating its information content, using Eq.~\ref{equ.shanno}.
The probability of a predicate is estimated by its frequency in an information source. An example of this can be shown in Fig.~\ref{fig:pred_dist}, where predicate ``on'' has higher occurrence frequency but fewer information contents when compared with ``standing on''. 
Based on a specific information source (\eg, Visual Genome), the frequency for each predicate category is used to calculate the information content. Next, we sort all predicates by their information content, resulting in ascending order. Top-$M$ predicates are chosen as common ones, and the rest are informative predicates. Next, we create a target domain by sampling labeled data from the source domain. A direct way is the balance sampling strategy that increases the sampling frequency of rare categories. However, such upsampling strategy is ineffective for the scene graph generation task, as demonstrated in previous work~\cite{scenegraph:ubtraing}.
Therefore, we take a separation undersampling strategy to create a target domain. In particular, we move all samples that belong to the informative predicate category in the source domain to the target domain. As for categories belonging to $M$ common predicates, we randomly sample $N$ labeled relationship triplet samples from source domains for each predicate, resulting in a balanced sample space between common predicates and informative ones. Besides balancing the dataset, randomly removing samples of common predicates may also have another benefit: because there are many ambiguous samples in common predicates that disturb informative predicates, as shown in Fig.~\ref{fig:conf_ann}, reducing samples of common predicates may help to reduce the impact of these samples to a certain extent. To eliminate the impact of ambiguous samples further, we also introduce another undersampling method, which uses the pre-trained SGG model in the first stage to find ambiguous samples of common predicates and remove them in the target domain. Details and Experiments can be found in Sec.~\ref{sec.ablation}.

After target domain construction, we can train an SGG model on it. Instead of training an SGG model from scratch on the target domain, we first initialize an SGG model from the one pretrained in the first stage. And then, only the last recognition layer of the SGG model is finetuned on the target domain. There are two reasons behind this: First, finetuning the whole model requires high computation costs. Second, training the whole model on the target domain with fewer labeled samples increases the risk of the overfitting problem, leading to the failure of recognizing common predicates. More details can be seen in Sec.~\ref{sec.ablation}.

\section{Experiments}

\subsection{Experimental Settings}
\noindent\textbf{Datasets.} Following previous work~\cite{scenegraph:KERN,scenegraph:motifs,scenegraph:ubtraing}, we conduct experiments on a widely used SGG dataset, namely Visual Genome (VG)~\cite{scenegraph:visual_genome}. It is composed of $108k$ images across $75k$ object categories together with $37k$ predicate classes. Since $92\%$ of the predicates have few samples, we followed previous works~\cite{scenegraph:IMP} and adopted a new VG split, containing the most frequent $150$ object categories and $50$ predicate classes. Moreover, the VG dataset is split into a training set ($70\%$) and a test set ($30\%$), and we further sample a validation set ($5k$) from the training set for model validation. 

\noindent\textbf{Model Configuration.} In this work, we evaluate our method based on three baselines: \textbf{MotifNet}~\cite{scenegraph:motifs}, \textbf{VCTree}~\cite{scenegraph:treelstm} and \textbf{Transformer}~\cite{networks:transformer,scenegraph:sggcode,scenegraph:rtn,scenegraph:gps}. The numbers of inner layers in the object encoder and the edge encoder are set to $4$ and $2$, respectively. Other hyperparameters are identical to the setting in \textbf{Model Zoo}~\cite{scenegraph:sggcode}. All models share the same settings and the pre-trained detector as well.

\noindent\textbf{Metrics.} Following previous works~\cite{scenegraph:ubtraing,scenegraph:treelstm,scenegraph:KERN}, we evaluate the SGG method on three subtasks: 1) Predicate Classification (PredCls), 2) Scene Graph Classification (SGCls), and 3) Scene Graph Detection (SGDet). The PredCls takes ground truth bounding boxes and labels as inputs. The SGCls takes ground truth object bounding boxes as inputs, but without labels. The SGDet requires predicting relationships from scratch. In this work, we use three evaluation metrics: R@K, mR@K, and mRIC@K. In detail, R@K averages the recall for all samples, while mR@K averages the recall across predicate categories. R@K underestimates informative predicate categories but focuses on common predicates with rich samples. Due to the defect existing in R@K measurement reported in~\cite{scenegraph:ubtraing}, we mainly report the mR@K metric, where it averages R@K for all predicate categories. 

Moreover, we propose a new metric, mRIC@K, that measures the mean recall with information content. Specifically, for each predicate, its recall with information content, denoted RIC, is calculated by multiplying its recall with the predicate’s information content, which is calculated by Eq.~\ref{equ.shanno}. mRIC@K, finally, can be obtained by averaging the RIC values across all predicate categories. This metric reflects how much information a scene graph can provide. A higher score implies that the generated scene graph involves more informative content and vice versa. Moreover, we use two information sources to estimate information content for each predicate, \ie, Visual Genome and Wikipedia. We separately denote mRIC (VG) and mRIC (Wiki) as the information content from Visual Genome and Wikipedia. 

\begin{table*}[ht]
	\begin{center}
		\resizebox{0.9\textwidth}{!}{	
			\begin{tabular}{c|c|c|c|c|c|c|c|c|c}
				\hline
				\multirow{2}{*}{Method} & \multicolumn{3}{c|}{PredCls}                  & \multicolumn{3}{c|}{SGCls}                    & \multicolumn{3}{c}{SGDet}                    \\ \cline{2-10} 
				& mR@20         & mR@50         & mR@100        & mR@20         & mR@50         & mR@100        & mR@20         & mR@50         & mR@100        \\ \hline
				Transformer            & 12.4          & 16.0          & 17.5          & 7.7           & 9.6           & 10.2          & 5.3           & 7.3           & 8.8           \\ \hline
				Transformer+\textbf{BPL}      & 24.5$^{+12.1}$ & 29.4$^{+13.4}$ & 31.7$^{+14.2}$ & 14.1$^{+6.4}$ & 16.8$^{+7.2}$ & 17.8$^{+7.6}$ & 10.2$^{+4.9}$ & {13.2}$^{+5.9}$ & {15.4}$^{+6.6}$ \\ \hline
				Transformer+\textbf{BPL}+\textbf{SA}     & {26.7}$^{+14.3}$ & {31.9}$^{+15.9}$ & {34.2}$^{+16.7}$ & {15.7}$^{+8.0}$ & {18.5}$^{+8.9}$ & {19.4}$^{+9.2}$ & {11.4}$^{+6.1}$ & {14.8}$^{+7.5}$ & {17.1}$^{+8.3}$ \\ \hline
				MotifNet                & 11.5          & 14.6          & 15.8          & 6.5           & 8.0           & 8.5           & 4.1           & 5.5           & 6.8           \\ \hline 
				MotifNet+\textbf{BPL}           & {22.6}$^{+11.1}$ & {27.1}$^{+12.5}$ & {29.1}$^{+13.3}$ & {13.0}$^{+6.5}$ & {15.3}$^{+7.3}$ & {16.2}$^{+7.7}$ & {9.7}$^{+5.6}$  & {12.4}$^{+6.9}$ & {14.4}$^{+7.6}$ \\ \hline
				MotifNet+\textbf{BPL}+\textbf{SA}          & {24.8}$^{+13.3}$ & {29.7}$^{+15.1}$ & {31.7}$^{+15.9}$ & {14.0}$^{+7.5}$ & {16.5}$^{+8.5}$ & {17.5}$^{+9.0}$ & {10.7}$^{+6.6}$ & {13.5}$^{+8.0}$ & {15.6}$^{+8.8}$ \\ \hline
				VCTree               & 11.7          & 14.9          & 16.1          & 6.2           & 7.5           & 7.9           & 4.2           & 5.7           & 6.9           \\ \hline
				VCTree+\textbf{BPL}           & {23.8}$^{+12.1}$ & {28.4}$^{+13.5}$ & {30.4}$^{+14.3}$ & {15.6}$^{+9.4}$ & {18.4}$^{+10.9}$ & {19.5}$^{+11.6}$ & {9.9}$^{+5.7}$  & {12.5}$^{+6.8}$ & {14.4}$^{+7.5}$ \\ \hline
				VCTree+\textbf{BPL}+\textbf{SA}          & {26.2}$^{+14.5}$ & {30.6}$^{+15.7}$ & {32.6}$^{+16.5}$ & {17.2}$^{+11.0}$ & {20.1}$^{+12.6}$ & {21.2}$^{+13.3}$ & {10.6}$^{+6.4}$ & {13.5}$^{+7.8}$ & {15.7}$^{+8.8}$ \\  
				\hline
			\end{tabular}
	
		}
	\vspace{-0.5em}
	\end{center}
	\caption{Ablation studies on generalizability and effectiveness of our proposed components, \ie, Balanced Predicate Learning (BPL) and Semantic Adjustment (SA). The proposed method is applied on three baseline models, where the MotifNet and VCTree are reimplemented by~\cite{scenegraph:ubtraing}. }
	
	\label{tab.eff_g2sct}
		\vspace{-1.5em}
\end{table*}

\begin{table}[h]
	\begin{center}	
		\resizebox{0.95\linewidth}{!}{
			\begin{tabular}{c|c|c|c|c}
				\hline
				Metric         & Method               & mRIC@20 & mRIC@50 & mRIC@100 \\ \hline
				\multirow{3}{*}{mRIC (VG)}        
				& Transformer      & 43.5  & 59.2  & 65.8  \\ \cline{2-5} 
				& Transformer+BPL  & 118.2  & 142.6 & 154.5  \\ \cline{2-5} 
				& Transformer+BPL+SA & 134.5  & 160.3 & 172.5  \\ \hline
				\multirow{3}{*}{mRIC (Wiki)} 
				& Transformer  & 74.1  & 96.6 & 105.9  \\ \cline{2-5} 
				& Transformer+BPL  & 182.1  & 216.3 & 231.3  \\ \cline{2-5} 
				& Transformer+BPL+SA & 204.6 & 239.2 & 254.4  \\ \hline
			\end{tabular}
		}
	\end{center}
\vspace{-0.5em}
	\caption{Information content assessment on the PredCls task. Models are evaluated on two information sources (Visual Genome \& Wikipedia).}
	\label{tab.inforc_g2sct}
	\vspace{-1.0em}
\end{table}
\subsection{Implementation Details}
We utilize Faster R-CNN~\cite{obj_det:faster} with ResNeXt-101-FPN~\cite{img_class:resnet,obj_det:fpn,obj_det:resxnet} pre-trained by~\cite{scenegraph:ubtraing} to detect instances in images and freeze the weights during learning scene graph generation. For scene graph generation, we firstly train all SGG models on the source domain according to the recommended configuration~\cite{scenegraph:sggcode} for all tasks, including the learning rate and batch size. Then each model is finetuned on the target domain with the same configuration. We take the logits before the softmax layer as $\Pr( y_{ij}^g | o_i, o_j;\theta )$ in Eq.~\ref{equ.transform} for the sake of simplicity.
For constructing the target domain, $M$ predicates with the smallest information content are set as common predicates, where $M$ is $15$ in this paper. Moreover, the sampling number $N$ is $2k$. We use the information content only from Visual Genome to distinguish common predicates and informative ones unless otherwise stated.

\begin{table*}[h]
	\begin{center}	
		\resizebox{0.93\textwidth}{!}{
			\begin{tabular}{c|c|c|c|c|c|c|c|c|c}
				\hline
				\multirow{2}{*}{Target Domain} & \multicolumn{3}{c|}{SGG (PredCls)} & \multicolumn{3}{c|}{mRIC (VG)} & \multicolumn{3}{c}{mRIC (Wiki)} \\ \cline{2-10} 
				& mR@20     & mR@50     & mR@100     & mRIC@20   & mRIC@50   & mRIC@100   & mRIC@20    & mRIC@50    & mRIC@100   \\ \hline
				General                        
				& 12.4      & 16.0      & 17.5       
				& 43.5  & 59.2  & 65.8   
				& 74.1  & 96.6 & 105.9    \\ \hline
				Wikipedia                     
				& 21.3      & 26.0      & 28.0       
				& 99.5    & 121.9   & 131.5    
				& 166.1    & 198.4   & 211.7    \\ \hline
				Visual Genome                  
				& 26.7      & 31.9      & 34.2       
				& 134.5  & 160.3 & 172.5    
				& 204.6    & 239.2    & 254.4    \\ \hline

		\end{tabular}}
	\end{center}
\vspace{-0.5em}
	\caption{Ablation studies of target domain construction in BPL. }
	\label{tab.is_mr}
	\vspace{-1.5em}
\end{table*}

\begin{table}[ht]
	\begin{center}
		\resizebox{0.95\linewidth}{!}{
			\begin{tabular}{c|c|c|c|c|c}
				\hline
				Settings & $C^{\star}$  & finetune & mR@20 & mR@50 & mR@100 \\ \hline
				1 & - & - & 24.5  & 29.4  & 31.7   \\ \hline
				2 & Random Init & \checkmark & 25.1  & 29.5  & 31.3   \\ \hline
				3 & SA & \checkmark & 25.8  & 30.2  & 32.1   \\ \hline
				4 & SA & - & 26.7  & 31.9  & 34.2   \\ \hline
		\end{tabular}}
	\end{center}
\vspace{-0.5em}
	\caption{Ablation studies of transition matrix on PredCls.}
	\label{tab.cof_mat}
\vspace{-0.5em}
\end{table}

\begin{table}[h]
	\begin{center}	
		\resizebox{0.6\linewidth}{!}{
			\begin{tabular}{c|c|c|c}
				\hline
				$\alpha$   & mR@20 & mR@50 & mR@100 \\ \hline
				$\alpha$ = 0.0 & 20.5 & 25.4 & 27.3  \\ \hline
				$\alpha$ = 0.3 & 25.8 & 30.5 & 32.8  \\ \hline
				$\alpha$ = 0.6 & 26.4 & 31.6 & 33.9  \\ \hline
				$\alpha$ = 1.0 & 26.7      & 31.9      & 34.2  \\ \hline 
				
		\end{tabular}}
	\end{center}
\vspace{-0.5em}
	\caption{Ablation studies of $\alpha$ in Eq.~\ref{equ.final_transition} on PredCls. }
	\label{tab.ftalpha}
\vspace{-0.5em}
\end{table}

\begin{table}[h]
	\begin{center}	
		\resizebox{0.8\linewidth}{!}{
			\begin{tabular}{c|c|c|c}
				\hline
				Undersampling  & PredCls & SGCls & SGDet \\ \hline
				Random & 26.7      & 15.7      & 11.4  \\ \hline
				Pre-trained Model & 28.5 & 16.4 & 11.4  \\ \hline
		\end{tabular}}
	\end{center}
	\vspace{-0.5em}
	\caption{Ablation studies of undersampling methods on mR@20 of the three subtasks.}
	\label{tab.eliminate_amb}
	\vspace{-0.5em}
\end{table}

\begin{table}[h]
	\begin{center}	
		\resizebox{0.95\linewidth}{!}{
			\begin{tabular}{c|c|c|c|c|c}
				\hline
				Settings   &   pretrained  & f-backbone   & R@20 & R@50 & R@100 \\ \hline
				1 & - & - & 28.4 & 34.9 & 37.2  \\ \hline
				2 & \checkmark & \checkmark & 30.6 & 37.3 & 39.5  \\ \hline
				3 & \checkmark & -  & 49.0 & 55.7 & 57.6  \\ \hline
		\end{tabular}}
	\end{center}
\vspace{-0.5em}
	\caption{Ablation studies of training approaches on PredCls.}
	\label{tab.wo_init}
	\vspace{-1.5em}
\end{table}

\subsection{Exploration Study}

In this section, we investigate the superiority of the proposed method from three aspects: 1). performance of scene graph generation; 2). information content of predicate prediction; and 3). accuracy of informative predicates.

For scene graph generation, we investigate the generalization capability of the proposed method by adding it to three baseline models as mentioned above. The experimental results are summarized in Tab.~\ref{tab.eff_g2sct}. As can be seen, all baseline models are improved by a large margin for all metrics. In particular, Transformer trained with BPL learning strategy outperforms the Transformer trained in source domain by $12.1$. It is further improved ($24.5$ \textit{vs.} $26.7$) when applying SA to cast common predicates as informative ones. Similar gains are obtained when applying our method to MotifNet and VCTree, indicating the generalizability of the proposed method. 

\begin{figure}[t]
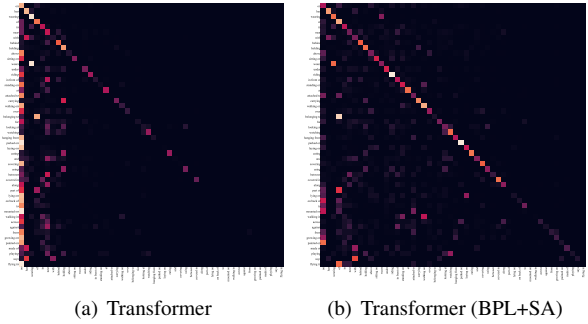

	\begin{center}	
		
		\subfigure[Transformer]
		{
			
			\includegraphics[width=0.45\linewidth]{./figure/conf_mat_transe.pdf}   
			
		}
		\subfigure[Transformer (BPL+SA)] 
		{
			
			\includegraphics[width=0.45\linewidth]{./figure/conf_mat_transe_ct.pdf}  
			
		}
	\vspace{-1.0em}
	\end{center}
	\caption{Confusion matrices of Transformer and Transformer (BPL+SA). The coordinates of these matrices are arranged in the increasing order of predicate information content from left to right and top to bottom.} 
	\label{fig:conf_mat_tg2s}
	\vspace{-1.5em}
\end{figure}
To evaluate whether the BA-SGG algorithm can provide more informative content, we respectively report evaluation results on mRIC@K for Transformer and the proposed method. The evaluation results based on two information sources are summarized in Tab.~\ref{tab.inforc_g2sct}. From the results, we can observe that the proposed methods obtain higher scores when compared with Transformer, which implies that the proposed method generates more informative scene graphs. 

In order to intuitively show the accuracy of informative predicates, we visualize confusion matrices across predicate categories. In particular, we compare two confusion matrices generated from Transformer and our method in Fig.~\ref{fig:conf_mat_tg2s}. Not surprisingly, the proposed method accurately detects more informative predicates than Transformer's since the confusion matrix generated from the proposed method is brighter on the diagonal than that of Transformer. Besides, Transformer (BPL+SA) gets rid of common predicates because its bright elements no longer concentrate on the left, compared with the confusion matrix of Transformer.

\subsection{Ablation Study on Details of Proposed Method}
\label{sec.ablation}

Except for the effectiveness of our method, we also investigate variants of our method for more insights. Next, we mainly study two variants of the proposed method: variants to SA and variants to BPL. Furthermore, we use Transformer as the baseline in the following experiments.

\noindent\textbf{Variants to SA.} For SA, we investigate variants of the transition matrix, \ie, $C^\star$ in Eq.~\ref{equ.final_transition}. In particular, we compare four settings:
1) Transformer model with BPL but without SA, where it is the baseline.
2) $C^\star$ is implemented by a random matrix and updated during training.
3) $C^\star$ is calculated by Eq.~\ref{equ.final_transition} and updated during training.
4) $C^\star$ is obtained by Eq.~\ref{equ.final_transition} while it is fixed during training.
The experimental results are shown in Tab.~\ref{tab.cof_mat}. From the results, we have the following observations: first, the transition matrix initialized from a random matrix and updated during training brings minor gains in terms of mean recall score. Second, the fixed transition matrix obtained from prediction priors (\ie, Eq.~\ref{equ.final_transition}) performs the best. This implies that updating the transition matrix during the training hurts the quality of the common-to-informative process. 

In Eq.~\ref{equ.final_transition}, we add an identity matrix and introduce a hyper-parameter $\alpha$ to investigate the importance of the identity matrix. The adjustment of $\alpha$ is shown in Tab.~\ref{tab.ftalpha}. From Tab.~\ref{tab.ftalpha}, we can find that the model not only needs to add the identity matrix but also needs to set $\alpha$ equal to $1$ to get better results.

\noindent\textbf{Variants to BPL.} As illustrated in Sec.~\ref{sec.pdtl}, the proposed BPL learning process consists of three steps, where the construction of the target domain is the essential part. To create the target domain, an information source is needed to calculate information content (IC) for each predicate category. Based on their IC values, predicates are further categorized as common or informative. Here, we use two different corpus sources for IC calculation: 1) Visual Genome, where it provides triplets for visual contents; and 2) Wikipedia, where it provides the latest articles (about $110k$).
Notably, Wikipedia is a language-based corpus source without visual contents. Next, two target domains can be generated based on these two information sources. Then, we perform diagnostic experiments to investigate whether the SSG model can benefit from the target domain with a balanced sample space. In particular, we compare three settings: 
1) Transformer model trained on the source domain.
2) Transformer model trained on target domain based on Wikipedia.
3) Transformer model trained on target domain based on Visual Genome.
The experimental results are summarized in Tab.~\ref{tab.is_mr}. We have the following findings: first, SGG models trained on the target domain, either language-based or visual-based, outperform the one trained on the source domain by a large margin (at least $8.9$ gains for mR@20). This indicates that the balanced sample space plays an important role in training the SGG model. Besides, the SGG model is improved after being trained on the language-based target domain, demonstrating the effectiveness and generalizability of the proposed IC-based predicate partition. As for the assessment of information contents, models trained on the target domain achieve higher scores than the model trained on source domain, either based on VG information source ($134.5$ \textit{vs.} $43.5$ and $99.5$ \textit{vs.} $43.5$) or Wiki information source ($204.6$ \textit{vs.} $74.1$, and $166.1$ \textit{vs.} $74.1$). This demonstrates that the scene graph generated from the models trained on the target domain provides more informative content for better scene understanding.

\begin{table*}[ht]
	\begin{center}	
		\resizebox{0.90\textwidth}{!}{
			\begin{tabular}{c|c|c|c|c|c|c|c|c|c}
				\hline
				\multirow{2}{*}{Method} & \multicolumn{3}{c|}{PredCls} & \multicolumn{3}{c|}{SGCls} & \multicolumn{3}{c}{SGDet}    \\ \cline{2-10} 
				& mR@20         & mR@50         & mR@100        & mR@20          & mR@50          & mR@100        & mR@20         & mR@50         & mR@100        \\ \hline
				IMP+~\cite{scenegraph:IMP,scenegraph:KERN}                    & -             & 9.8           & 10.5          & -              & 5.8            & 6.0           & -             & 3.8           & 4.8           \\ \hline
				FREQ~\cite{scenegraph:motifs,scenegraph:treelstm}                    & 8.3           & 13.0          & 16.0          & 5.1            & 7.2            & 8.5           & 4.5           & 6.1           & 7.1           \\ \hline
				KERN~\cite{scenegraph:KERN}                    & -             & 17.7          & 19.2          & -              & 9.4            & 10.0          & -             & 6.4           & 7.3           \\ \hline
				GPS-Net~\cite{scenegraph:gps}                 & -             & -             & 22.8          & -              & -              & 12.6          & -             & -             & 9.8           \\ \hline
				GB-Net~\cite{scenegraph:gbnet}     & - & 22.1 & 24.0 & -  & 12.7  & 13.4 & - & 7.1 & 8.5 \\ \hline
				
				MotifNet~(Focal)~\cite{scenegraph:motifs,scenegraph:ubtraing}           & 10.9          & 13.9          & 15.0          & 6.3            & 7.7            & 8.3           & 3.9           & 5.3           & 6.6           \\ \hline
				MotifNet~(Reweight)~\cite{scenegraph:motifs,scenegraph:ubtraing}        & 16.0          & 20.0          & 21.9          & 8.4            & 10.1           & 10.9          & 6.5           & 8.4           & 9.8           \\ \hline
				MotifNet~(Resample)~\cite{scenegraph:motifs,scenegraph:ubtraing}        & 14.7          & 18.5          & 20.0          & 9.1            & 11.0           & 11.8          & 5.9           & 8.4           & 9.8           \\ \hline
				MotifNet~(TDE)~\cite{scenegraph:motifs,scenegraph:ubtraing}             & 18.5          & 24.9          & 28.3          & 11.1           & 13.9           & 15.2          & 6.6           & 8.5           & 9.9           \\ \hline
				
				VCTree~(TDE)~\cite{scenegraph:treelstm,scenegraph:ubtraing}             & 18.4          & 25.4          & 28.7          & 8.9            & 12.2           & 14.0          & 6.9           & 9.3           & 11.1          \\ \hline
				VTransE~(TDE)~\cite{scenegraph:VTEN,scenegraph:ubtraing}            & 18.9          & 25.3          & 28.4          & 9.8            & 13.1           & 14.7          & 6.0           & 8.5           & 10.2          \\ \hline 
				\hline
				
				Transformer~(BA-SGG)     & \textbf{26.7} & \textbf{31.9} & \textbf{34.2} & \textbf{15.7} & \textbf{18.5} & \textbf{19.4} & \textbf{11.4} & \textbf{14.8} & \textbf{17.1} \\ \hline
				MotifNet~(BA-SGG)          & \textbf{24.8} & \textbf{29.7} & \textbf{31.7} & \textbf{14.0} & \textbf{16.5} & \textbf{17.5} & \textbf{10.7} & \textbf{13.5} & \textbf{15.6} \\ \hline
				VCTree~(BA-SGG)          & \textbf{26.2} & \textbf{30.6} & \textbf{32.6} & \textbf{17.2} & \textbf{20.1} & \textbf{21.2} & \textbf{10.6} & \textbf{13.5} & \textbf{15.7} \\ \hline
		\end{tabular}}
	\end{center}
\vspace{-0.5em}
	\caption{Comparison between our method (BA-SGG) and previous methods.}
	\label{tab.compare}
	\vspace{-0.5em}	
\end{table*}
\begin{figure*}[ht]
	\begin{center}
		\includegraphics[width=0.9\linewidth]{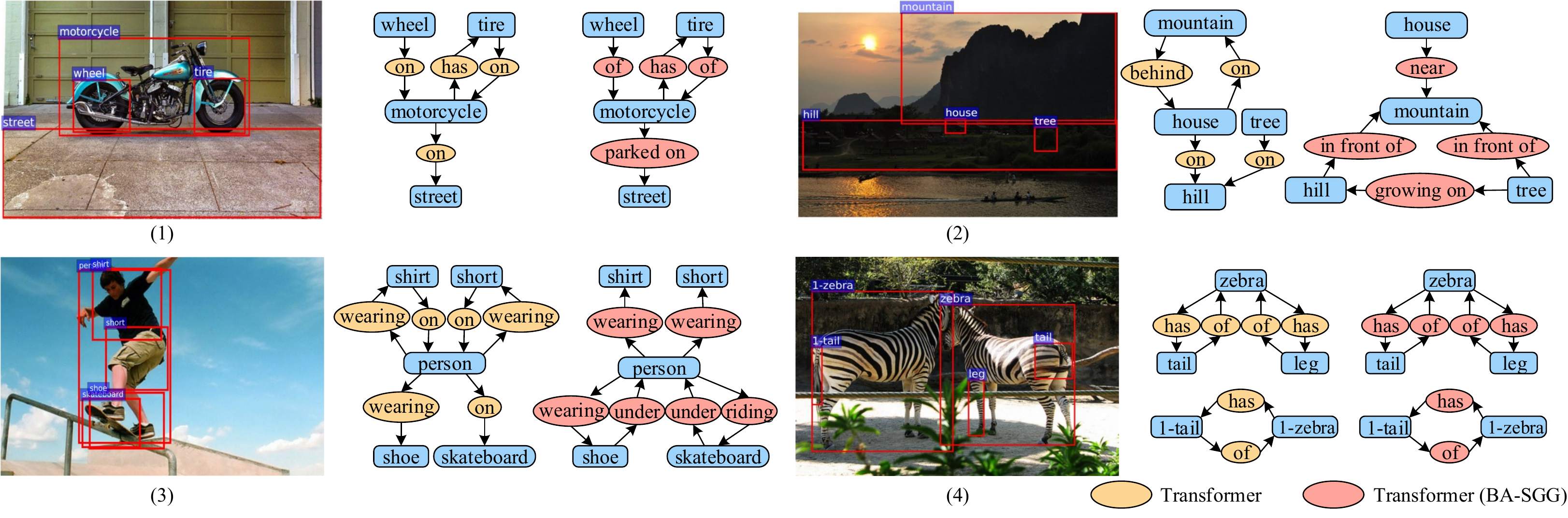}
	\end{center}
\vspace{-1.5em}	
	\caption{Visualization results of Transformer and Transformer (BA-SGG) on the PredCls task. The generated scene graph from the Transformer (BA-SGG) is more informative than the one from Transformer. Only top 30\% relationships on each image are shown for clarity.}
	\label{fig:vis_res}
\vspace{-1.5em}	
\end{figure*}

To further eliminate the impact of ambiguous samples in common predicates, as shown in Fig.~\ref{fig:conf_ann}, we design another undersampling method. Intuitively, the ambiguous sample in Fig.~\ref{fig:conf_ann} can be annotated as either a common predicate ``on'' or an informative predicate ``standing on''. In other words, the confidence of the annotation ``on'' should be less. Therefore, we use the confidence of the pre-trained SGG model in the first stage to remove ambiguous samples. Specifically, we input all training samples to the model and keep the top-2k samples with the highest confidence for each common predicate in the target domain. As shown in Tab.~\ref{tab.eliminate_amb}, although using the pre-trained model for undersampling improves the performances on PredCls, this improvement does not achieve the desired effect, especially on SGDet. There may be two reasons for the insignificant improvement: 1). Random undersampling by a large margin (\eg, ``on'' from $70k$ to $2k$) can reduce the influence of ambiguous samples to a great extent. 2). Using the pre-trained model to find ambiguous samples is not very reliable, and another effective method should be introduced. Therefore, this problem should still be valued in future work, and we use the random undersampling method in this paper for simplicity and effectiveness.

After the target domain construction, a good training approach is needed. To explore this aspect, we perform an additional experiment as illustrated in Tab.~\ref{tab.wo_init}, where it compares three different training settings:
1) Training an SGG model from scratch on the target domain.
2) Finetuning a whole SGG model on the target domain, where it is pretrained on the source domain. 
3) Finetuning the classifier layer of an SGG model on the target domain, where it is pretrained on the source domain.
From Tab.~\ref{tab.wo_init}, we can observe that the third setting provides the best result, whilst directly training an SGG model from scratch on the adjusted domain gives the worst
result. Furthermore, the model trained with the third setting outperforms the one trained with the second setting. One possible reason is that the scale of sample space in the target domain (about $55k$) is much smaller than the source domain's (about $400k$), posing high risks of overfitting. Since this experiment focuses on the training approach, we remove SA in this experiment.

\subsection{Comparison with State-of-the-art Methods}
After verifying the effect of the proposed method, we compare it with state-of-the-art methods. The comparison results are shown in Tab.~\ref{tab.compare}. As shown in the results, the proposed method achieves the best performance in all the measure metrics among all the comparison methods, reaching $26.7$ mR@20 for PredCls, $15.7$ mR@20 for SGCls, and $11.4$ mR@20 for SGDet, respectively. For PredCls, it outperforms the best competitor TDE by $6.3$ for MotifNet and $7.8$ for VCTree on mR@20. Similar gains also are observed for another two tasks. Compared with the resample or reweight methods, \eg, MotifNet~(Focal), MotifNet~(Reweight), and MotifNet~(Resample), our method also outperforms them because they do not simultaneously consider the two imbalances, \ie, semantic space level imbalance and training sample level imbalance. This clearly indicates the superiority of our method and the importance of the two imbalances.

\subsection{Visualization Results}
Qualitative results across various scenes are shown in Fig.~\ref{fig:vis_res}. Here, we compare Transformer with Transformer equipped with the proposed method (BA-SGG). From Fig.~\ref{fig:vis_res}, Transformer (BA-SGG) generates more informative scene graphs than vanilla Transformer, such as (\{motorcycle, parked on, street\} \textit{vs.} \{motorcycle, on, street\}) in the first example,  (\{tree, growing on, hill\} \textit{vs.} \{tree, on, hill\}) in the second example, and (\{person, riding, skateboard\} \textit{vs.} \{person, on, skateboard\}) in the third example. This clearly demonstrates the success of predicates adjustment by the proposed method. However, the fourth example shows both models fail to recognize the predicate between two zebras since both zebras can be viewed as subjects. This implies that recognition of the subject-subject interaction relationship is still a bottleneck for SGG.

\section{Conclusion}
In this work, balance adjustment scene graph generation (BA-SGG) is proposed for learning scene graphs from general to specific. We first reveal the imbalance between common predicates and informative predicates. Then two strategies are designed: Semantic Adjustment and Balanced Predicate Learning, in view of the imbalance. The semantic adjustment strategy explores the semantic relation between predicates, and the balanced predicate learning strategy transfers the knowledge learned from general predicates to informative ones. Finally, numerous experiments show that our method significantly improves the performance of scene graph generation.

\noindent\textbf{Acknowledgement.} This work is funded by Kuaishou.

{\small
\bibliographystyle{ieee_fullname}
\balance
\bibliography{egbib}
}

\end{document}